\documentclass[letterpaper, 10 pt, conference]{ieeeconf}  
\usepackage{graphicx}
\usepackage{amsmath}
\usepackage{amssymb}
\usepackage{placeins}
\usepackage{subcaption} 
\usepackage{caption}
\usepackage{xcolor}
\usepackage{hyperref}
\usepackage{booktabs}
\usepackage{cite}
\usepackage{makecell} 

\usepackage{multirow}
\usepackage{tabularx}
\usepackage{algorithm}
\usepackage{algorithmicx}
\usepackage{algpseudocode}

\IEEEoverridecommandlockouts                              
\title{\LARGE \bf
Exploring Transformer-Augmented LSTM for Temporal and Spatial Feature Learning in Trajectory Prediction
}

\author{Chandra Raskoti and Weizi Li
\thanks{Chandra Raskoti and Weizi Li are with the Min H. Kao Department of Electrical Engineering and Computer Science at the University of Tennessee, Knoxville, TN, USA {\tt\small craskoti@vols.utk.edu, weizili@utk.edu}}%
\thanks{}
}

\begin{document}

\maketitle
\thispagestyle{empty}
\pagestyle{empty}

\begin{abstract}
Accurate vehicle trajectory prediction is crucial for ensuring safe and efficient autonomous driving. This work explores the integration of Transformer based model with Long Short-Term Memory (LSTM) based technique to enhance spatial and temporal feature learning in vehicle trajectory prediction. Here, a hybrid model that combines LSTMs for temporal encoding with a Transformer encoder for capturing complex interactions between vehicles is proposed. Spatial trajectory features of the neighboring vehicles are processed and goes through a masked scatter mechanism in a grid based environment, which is then combined with temporal trajectory of the vehicles. This combined trajectory data are learned by sequential LSTM encoding and Transformer based attention layers. The proposed model is benchmarked against predecessor LSTM based methods, including STA-LSTM, SA-LSTM, CS-LSTM, and NaiveLSTM. Our results, while not outperforming it's predecessor, demonstrate the potential of integrating Transformers with LSTM based technique to build interpretable trajectory prediction model. Future work will explore alternative architectures using Transformer applications to further enhance performance. This study provides a promising direction for improving trajectory prediction models by leveraging transformer based architectures, paving the way for more robust and interpretable vehicle trajectory prediction system.

\end{abstract}


\section{Introduction}
\label{sec:introduction}

Since the first autonomous driving competition hosted by DARPA in 2005, autonomous vehicles (AVs) have gained significant attention from academia and industry. With advancements in sensing technologies and machine learning, the development of autonomous driving systems has achieved remarkable progress. One critical component of autonomous driving is vehicle trajectory prediction, which ensures the safe and efficient navigation of AVs by predicting the future motions of surrounding vehicles \cite{c1}.

There are two primary approaches to achieving autonomous driving: the end-to-end approach, which directly maps raw sensor data to control commands using a unified model \cite{c2}, and the traditional engineering approach, which relies on modular systems for tasks such as detection, tracking, prediction, and planning \cite{c3}. The traditional engineering approach is often favored for its better interpretability and controllability, particularly in safety-critical systems such as AVs.

Among the techniques for trajectory prediction, recurrent neural networks (RNNs) and their variants, such as Long Short-Term Memory (LSTM) networks, have been widely employed due to their ability to capture temporal dependencies in sequential data \cite{c4}. The STA-LSTM model, which integrates spatiotemporal attention mechanisms, represents a notable advancement in this field by addressing the challenge of model explainability. Through temporal and spatial attention mechanisms, STA-LSTM not only predicts trajectories with high accuracy but also provides insights into the influence of historical and neighboring vehicle trajectories \cite{c5}.

Transformer models have emerged as a promising alternative due to their ability to capture long-range dependencies through self-attention mechanisms \cite{c6}. Unlike sequential models such as LSTMs, Transformers process entire sequences simultaneously, allowing them to better capture both global and local interactions. This characteristic makes them particularly well-suited for vehicle trajectory prediction, where understanding complex interactions between the target and neighboring vehicles is crucial \cite{c7}.

In this study, we extend the STA-LSTM framework by integrating a vanilla Transformer architecture to explore its potential for enhancing trajectory prediction. Our work maintains the core methodology of STA-LSTM while replacing its sequential processing with the parallelized attention mechanisms of Transformers. By evaluating the performance of the Transformer-enhanced model against the original STA-LSTM, we aim to assess the feasibility of employing Transformers for trajectory prediction and analyze their impact on model explainability and accuracy.

This paper is structured as follows. Section~\ref{sec:related} reviews related work in trajectory prediction based on traditional, deep learning based LSTM  methods and then transformer based techniques. Section~\ref{sec:method} presents our methodology, detailing the integration of Transformer architectures with the existing LSTM based framework. Similarly experimental setup and how experiments is proceeded is discussed in section~\ref{sec:exp}. Then, section~\ref{sec:results} provides a comparative analysis of experimental results. Finally, section~\ref{sec:conclude} discusses the implications of our findings and future research directions. 
\section{Related Work}
\label{sec:related}

\subsection{Traditional Methods for Trajectory Prediction}
Conventional approaches to vehicle trajectory prediction include physics based, maneuver based, and interaction-aware models. Physics based methods rely on kinematic and dynamic constraints such as acceleration, yaw rate, and road friction to predict short-term trajectories. While accurate for minimal time horizons, these models often fail to account for complex driver behaviors or interactions with surrounding vehicles \cite{c8}. Maneuver based models enhance physics based methods by incorporating predefined driver intentions, such as lane changes or turns, to improve prediction accuracy as in \cite{c9}. However, both approaches are limited in capturing interactions among vehicles, a critical factor in crowded or dynamic traffic environments.

\subsection{Deep Learning Approaches with LSTMs}
The advent of deep learning has revolutionized trajectory prediction by enabling data-driven modeling of complex temporal and spatial dependencies. Long Short-Term Memory (LSTM) networks, a class of recurrent neural networks, have been widely adopted due to their ability to model long-term dependencies in sequential data.

For example, Deo and Trivedi \cite{c4} proposed a convolutional social pooling LSTM model that considers vehicle interactions for trajectory prediction by aggregating spatial features from surrounding vehicles. This approach demonstrated the potential of LSTMs in managing traffic dynamics through multi-agent interactions.

To enhance context-awareness, STA-LSTM integrates spatiotemporal attention mechanisms, allowing the model to assign attention weights to both historical trajectories and neighboring vehicles. This enhances interpretability and provides valuable insights into the factors influencing the target vehicle’s future motion. The spatial attention mechanism focuses on nearby vehicles, while temporal attention tracks the influence of past trajectory points.

Further advancements in LSTM based trajectory models include the Social GAN proposed by Gupta et al. \cite{c13}, which models human-like interactions using LSTM based encoders and decoders combined with generative adversarial networks. Similarly, Sadeghian et al. \cite{c14} introduced Sophie, a socially-aware trajectory prediction framework that leverages LSTMs with attention mechanisms and semantic scene contexts for better prediction in complex environments.

More recently, Amirian et al. \cite{c15} extended LSTM based methods by integrating variational autoencoders (VAEs) to model trajectory uncertainty, demonstrating the flexibility of LSTMs in probabilistic forecasting. Such models enable dynamic uncertainty estimation in real-world scenarios where trajectory outcomes are inherently uncertain due to unpredictable human behavior.

These works underscore the continued relevance and adaptability of LSTM based models in trajectory prediction, especially when integrated with attention mechanisms, social pooling, and probabilistic components.

\subsection{Transformer Models for Trajectory Prediction}
Transformer models have emerged as a groundbreaking approach in time-series modeling due to their ability to capture long-range dependencies through self-attention mechanisms \cite{c6}. Unlike LSTMs, which process sequences sequentially, Transformers evaluate all elements of a sequence simultaneously, enabling efficient modeling of complex vehicle interactions in traffic environments. This characteristic makes Transformers well-suited for tasks involving simultaneous spatial and temporal reasoning.

The self-attention mechanism in Transformers dynamically assigns importance weights across the entire sequence, allowing the model to focus on key features at different time steps. This capability makes Transformers advantageous for modeling complex vehicle-vehicle and vehicle-environment interactions, where dependencies can occur across non-adjacent time steps \cite{c8}.

Zhou et al. \cite{c7} demonstrated the potential of Transformers in long-sequence time-series forecasting, highlighting their ability to integrate both global and local feature patterns. Their work emphasized how self-attention enables the effective aggregation of contextual information across extended prediction horizons. Similarly, Liu et al. \cite{c11} showed that Transformers outperform traditional RNNs in capturing intricate temporal relationships in dynamic systems.

In trajectory prediction, where interaction-aware modeling is crucial, applying Transformer architectures can provide a robust framework for predicting multi-agent trajectories. Recent work by Chen et al. \cite{c12} illustrated how attention based architectures improved trajectory forecasting by learning high-level spatiotemporal interactions, paving the way for more interpretable and accurate autonomous driving models.
\section{Methodology}
\label{sec:method}

\subsection{Data Preprocessing and Description}
The NGSIM dataset includes vehicle trajectory data collected from the US-101 and I-80 freeways, providing detailed information such as dataset and vehicle IDs, frame numbers, and positional data local X and Y coordinates. Data preprocessing includes extracting essential fields such as vehicle and frame IDs, and standardizing lane IDs by capping the maximum value at six for consistency. Maneuver detection involves calculating lateral and longitudinal movements using look-back and look ahead time windows of 40 frames (~4 seconds). Neighboring vehicles are identified by defining a spatial grid of 13 longitudinal cells per lane, covering a range of -90 to +90 meters, with grid indexing based on relative positions. Each vehicle's closest neighbors are assigned to corresponding cells in the grid. Edge cases where vehicles lack sufficient trajectory history or future data are removed to ensure model reliability. Finally, the dataset is divided into training (70\%), validation (10\%), and test (20\%) sets, ensuring unique vehicle IDs across split. As a result, the training, validation, and test data set have 5,922,867, 859,769, and 1,505,756 entries, respectively. No extra preprocessing such as normalization is applied to the data set.

\subsection{Temporal Representation Learning}
At time step $t$, vehicle's $T$ historical trajectory ${X_{t-T+1}^{\text{v}}, ..., X_t^{\text{v}}}$ is embedded and passed through an LSTM encoder, producing sequential hidden states. 
\begin{equation}
H_t = \text{LSTMEnc}(\{X_{t-T+1}, ..., X_t\})
\end{equation}
These hidden states are then processed using a Transformer encoder layer to capture temporal dependencies using multi-head attention and a feed-forward layer. This mechanism enhances the model's capability to consider both short-term and long-term temporal dependencies.

\begin{equation}
Z_t = \text{TransEncoder}(H_t), \mid H_t \in \mathbb{R}^{T \times d}
\end{equation}
Where, $d$ is the embedding size for LSTM encoder.

\subsection{Spatial Representation Learning}
At time step $t$, the $T$-step historical trajectory ${G_{t-T+1}^v, ..., X_t^v}$ of vehicle $v$ which can be either the target vehicle or a neighboring vehicle is taken as input to an LSTM encoder. 
\begin{equation}
S_t = \text{LSTMEnc}(\{G_{t-T+1}, ..., G_t\})
\end{equation}

These embeddings are further passed through a Transformer encoder layer with eight attention heads and a feed-forward network. 

\begin{equation}
Z_t^{\text{nbrs}} = \text{TransEncoder}(S_t), \mid S_t \in \mathbb{R}^{T \times d}
\end{equation}
A masked scatter operation assigns neighbor encodings into grid cells based on their relative positions, forming a structured spatial context representation. 
\begin{equation}
\text{soc\_enc} = \text{MaskedScatter}(M, Z_t^{\text{nbrs}}),
\end{equation}
\begin{equation}
\quad M \in \{0, 1\}^{B \times C \times G \times d}
\end{equation}

Where:
\begin{itemize}
    \item $M$ is the binary mask indicating vehicle presence
    \item $B$ is the batch size
    \item $C$ is the number of channels (neighbor directions)
    \item $G$ is the grid size
    \item $d$ is the embedding dimension
\end{itemize}

This structured representation captures dynamic spatial relationships critical for predicting interactions in dense traffic environments.

\subsection{Decoder Module}

The decoder in our proposed model uses an LSTM-layers for predicting vehicle trajectories. The combined feature representation from both spatial and temporal encodings serves as input to the decoder. This feature vector is processed by an LSTM decoder with a hidden state size of 128, whereas the encoded input features from the Transformer and LSTM encoders have a dimensionality of 64.
After the LSTM processes the feature representation, its output is passed through a fully connected linear layer. This layer has two output neurons representing the predicted $x$ and $y$ coordinates of the vehicle's position. The decoder produces predictions for 5 future time steps, generating a sequence of 2D coordinates representing the vehicle’s future trajectory.

\begin{figure*}[h!]
\centering
\resizebox{\textwidth}{!}{\includegraphics{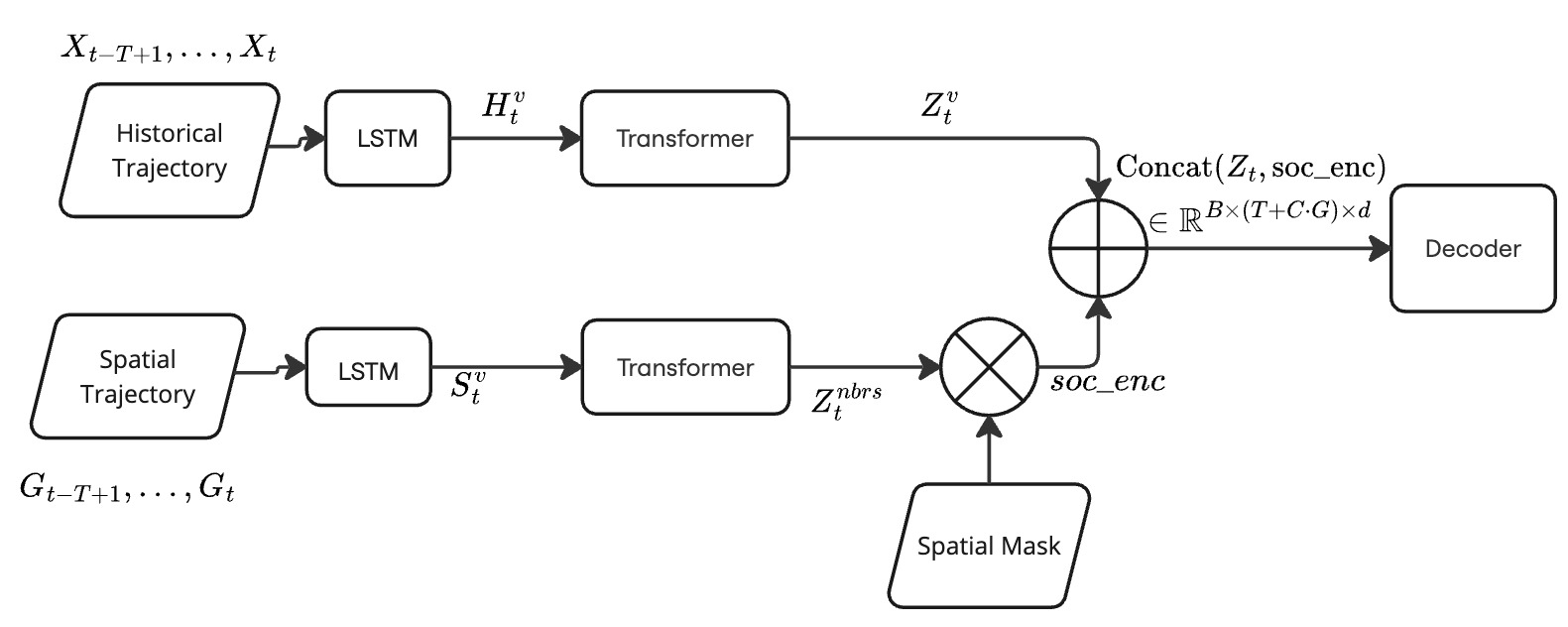}}
\caption{Architecture diagram illustrating the end-to-end encoding process of spatial and temporal features in our proposed method. The historical trajectories of the target vehicle and its neighboring vehicles are processed separately through LSTM and Transformer encoders to extract temporal and spatial features. Spatial features are further refined using a masked scatter mechanism based on vehicle presence. The resulting spatial and temporal embeddings are concatenated into a unified feature representation. This combined encoding is then passed through an LSTM based decoder to predict the future $\hat{x}, \hat{y}$ coordinates of the target vehicle over prediction time steps 1 through 5.}
\label{fig:Encoder architecutre}
\end{figure*}


\vspace{3cm}
\hrule
\vspace{0.2cm}
\textbf{Algorithm}
\vspace{0.2cm}
\hrule\vspace{0.05cm}\hrule
\vspace{0.2cm}
\textbf{Input:} Historical trajectories $\{X_{t-T+1}^v, ..., X_t^v\}$ for the target vehicle and its neighbors, binary mask $M$ indicating vehicle presence and other hyperparameters

\begin{enumerate}
    \item Initialize $\text{LSTMEnc}$, $\text{TransEncoder}$, and $\text{LSTMDec}$ with random weights

    \item \textbf{For each epoch} $e$ in $E$ \textbf{do}:
       - Extract historical trajectory data $X_t^v$ and neighboring vehicle data $G_t$
        \\
       - Encode target vehicle's historical trajectory using LSTM:
        \[ H_t = \text{LSTMEnc}(X_t^v) \]
       - Apply Transformer encoder on $H_t$:
        \[ Z_t = \text{TransEnc}(H_t) \]
        - Encode neighbors' historical trajectories using LSTM:
        \[ S_t = \text{LSTMEnc}(G_t) \]
        - Apply Transformer encoder on $S_t$:
        \[ Z_t^{\text{nbrs}} = \text{TransEnc}(S_t) \]
        - Apply mask $M$ to handle vehicle presence:
        \[ \text{soc\_enc} = \text{MaskedScatter}(M, Z_t^{\text{nbrs}}) \]
        - Concatenate temporal and spatial features:
        \[ \text{CombinedEnc} = \text{Concat}(Z_t, \text{soc\_enc}) \]
        - Decode concatenated features through LSTM based decoder:
        \[ \hat{Y}_t = \text{LSTMDec}(\text{CombinedEnc}) \]
        - Compute prediction euclidean distance loss between predicted and ground truth coordinates:
        \[ \mathcal{L} = \frac{1}{B} \sum_{i=1}^{B} \sum_{t=1}^{T} ||(\hat{X}_t,\hat{Y}_t) - (X_t,Y_t)||^2 \]
        $B$ : Batch Size
        $T$ : Number of prediction steps
        \\
        \\
        - Update $\text{LSTMEnc}$, $\text{TransEncoder}$, and $\text{LSTMDec}$ using Adam optimizer with learning rate $\alpha$

\end{enumerate}
\hrule
\vspace{0.2cm}
\textbf{Output:} Predicted future trajectories ($\hat{X}_t$, $\hat{Y}_t$) for the target vehicle
\vspace{0.1cm}
\hrule
\section{Experimental Setup}
\label{sec:exp}

To evaluate the proposed model, we follow the same data processing procedures as the STA-LSTM model. Each vehicle trajectory is downsampled by a factor of 2, and the coordinate space centered around the target vehicle is discretized into a $3 \times 13$ spatial grid, where the $y$-axis represents the highway's motion direction and $x$-axis represents the highway lane encoding. A time step of 0.2 seconds is used.

For data preparation, we considered 15-step (3-second) historical trajectories of the target and neighboring vehicles within the $3 \times 13$ grid as inputs. The task is to predict the next 5-step (1-second) future trajectory of the target vehicle. This process minimizes the following objective:

\begin{equation}
L(\theta) = \frac{1}{N} \sum_{i=1}^{N} \sum_{t=1}^{T} \left[ (\hat{x}_{i,t} - x_{i,t})^2 + (\hat{y}_{i,t} - y_{i,t})^2 \right] \\
\end{equation}
\renewcommand{\labelitemi}{}
\begin{itemize}
    \item $N$ = Number of training samples
    \item $T$ = Number of predicted time steps
    \item $(\hat{x}_{i,t}, \hat{y}_{i,t})$ = Predicted coordinates for sample $i$ at time step $t$
    \item $(x_{i,t}, y_{i,t})$ = Actual coordinates for sample $i$ at time step $t$
    \item $L(\theta)$ = MSE on Euclidean distance
\end{itemize}
\vspace{1cm}
For model configuration, the embedding size is set to 32, while the LSTM hidden state dimension is set to 64. The feed-forward layer in the Transformer encoder has a dimension of 512. We train the model using the Adam optimizer with a learning rate of 0.001. The model is trained for 50 epochs on an Intel i9-14900K CPU, an Nvidia GTX 4090 GPU, and 32GB RAM. The total training time is approximately 15 hours. These settings ensure a fair comparison with the STA-LSTM model while leveraging the extended feature representation provided by the Transformer-enhanced architecture.
\section{Results}
\label{sec:results}

We benchmark our per proposed method against previous LSTM based methods, including STA-LSTM, SA-LSTM, CS-LSTM, and NaiveLSTM. The comparison is conducted step-by-step from prediction steps 1 through 5, following the same evaluation setting described in the original paper \cite{c5}. Since these benchmark models were not implemented or retrained except for STA-LSTM, their performance metrics are taken from published results. Our experimental setup replicates the original study's conditions to enable a fair comparison. 

\begin{figure}[h!]
\centering
\includegraphics[width=\linewidth]{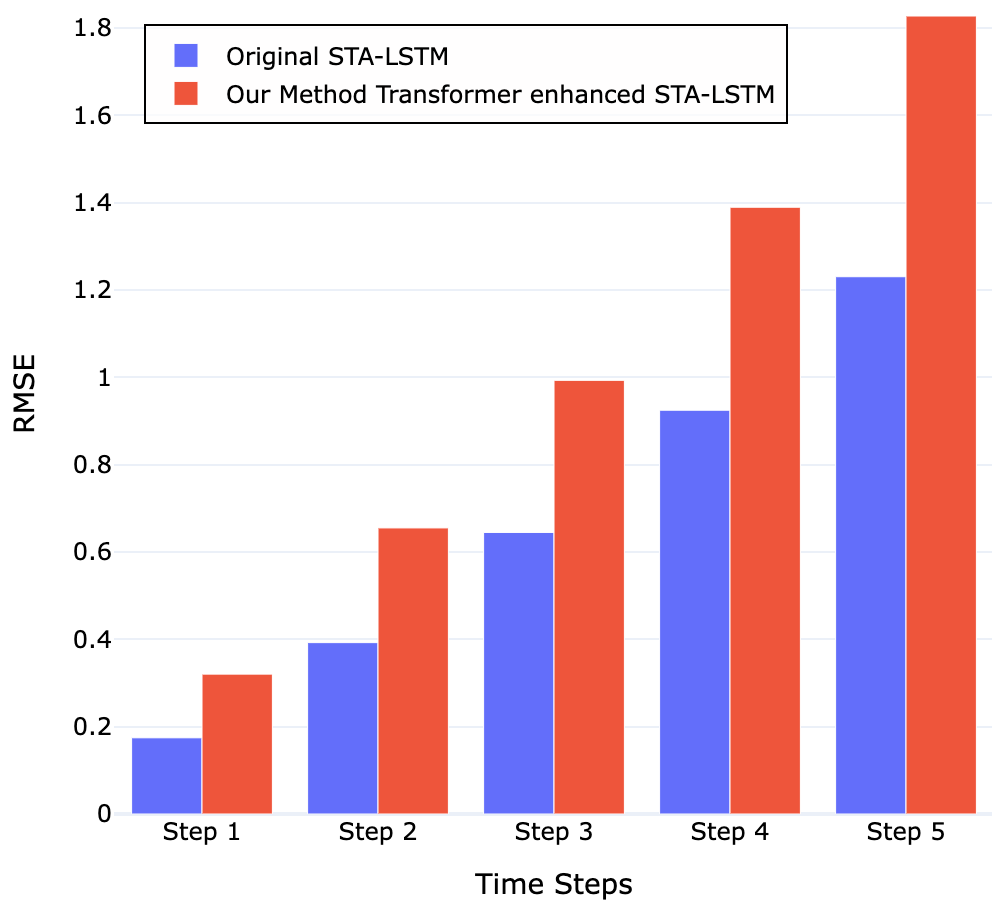}
\caption{RMSE comparison of two experiments performed: Original STA-LSTM method, and our Transformer-enhanced STA-LSTM. As expected, the RMSE increases across both models as prediction steps progress from 1 to 5.}
\label{fig:RMSE_comparison_plot}
\end{figure}

Table I shows the results of benchmark methods, our method, and the run we performed for STA-LSTM. A noticeable difference exists between the performance reported in the original paper and our run, requiring further investigation. Among all benchmark methods, our model performs the poorest, indicating the need for continuous improvements through future experiments. As expected, the RMSE increases across all models as prediction steps progress from 1 to 5. Our model similarly exhibits degraded performance as the prediction horizon lengthens, consistent with patterns observed in other benchmark models. This is further clear to visualize in Figure \ref{fig:RMSE_comparison_plot}. 
While our proposed model did not outperform the predecessor STA-LSTM method, the results still provides a promising basis for further exploration. The preliminary results suggest potential improvements by refining how the Transformer is integrated into the LSTM framework. Future work will focus on better applying the Transformer layer and optimizing the overall architecture, aiming to surpass state-of-the-art performance in vehicle trajectory prediction.

\begin{table}[h!]
\centering
\label{benchmark results}
\caption{A Comparison of our model with benchmark models using RMSE per Prediction Time Step (0.2s)}
\begin{tabular}{l c c c c c}
\hline
\textbf{Model} & \textbf{Step 1} & \textbf{Step 2} & \textbf{Step 3} & \textbf{Step 4} & \textbf{Step 5} \\ 
\hline
\hline
\makecell{Physics based model \\ result from paper\cite{c5}}& 0.1776 & 0.3852 & 0.6033 & 0.8377 & 1.0888  \\ 
\hline
\makecell{Naïve LSTM \\ result from paper\cite{c5}} & 0.1012 & 0.2093 & 0.3384 & 0.4830 & 0.6406  \\ 
\hline
\makecell{SA-LSTM \\ result from paper\cite{c5}} & 0.1026 & 0.2031 & 0.3157 & 0.4367 & 0.5643  \\ 
\hline
\makecell{CS-LSTM \\ result from paper\cite{c5}} & 0.1029 & 0.2023 & 0.3146 & 0.4364 & 0.5674  \\ 
\hline
\makecell{STA-LSTM \\ result from paper\cite{c5}} & 0.0995 & 0.2002 & 0.3130 & 0.4348 & 0.5615 \\ 
\hline
\makecell{STA-LSTM \\ result from our run} & 0.1745 & 0.3924 & 0.6447 & 0.9245 & 1.2302 \\
\hline
\makecell{Transformer\\enhanced STA-LSTM\\(our method)} & 0.3203 & 0.6551 & 0.9931 & 1.3893 & 1.8273 \\
\hline
\end{tabular}
\end{table}

\section{Conclusion and Future Work}
\label{sec:conclude}
In this work, we introduced a integration of Transformer and LSTM based architecture forr vehicle trajectory prediction, leveraging both spatial and temporal feature learning mechanisms. The proposed model processes spatial relationships through a grid based masked scatter mechanism while utilizing LSTMs and Transformer encoders to model temporal dynamics and long term dependencies. Our approach was benchmarked against established LSTM based methods, including STA-LSTM, SA-LSTM, CS-LSTM, and NaiveLSTM. 

While our proposed model did not surpass the performance of the state-of-the-art STA-LSTM model, our findings highlight the potential benefits of incorporating Transformer based self-attention into trajectory prediction frameworks. The preliminary results suggest that further exploration of architectural design, attention mechanisms, and integration strategies may yield improved predictive accuracy. This study demonstrates that combining LSTMs with Transformer based learning technique can be a a promising step toward more accurate and interpretable autonomous driving systems.


There are many future directions that we can pursue. 
First, we would like to integrate this project with existing techniques~\cite{Li2019ADAPS,Shen2022IRL,Shen2021Corruption,Villarreal2024AutoJoin} to further enable the learning, planning, and control of robot vehicles. 
Second, we want to extend our study to large-scale traffic simulation, reconstruction, and prediction~\cite{Wilkie2015Virtual,Li2017CityFlowRecon,Li2017CitySparseITSM,Li2018CityEstIET,Lin2019Compress,Lin2019BikeTRB,Chao2020Survey,Poudel2021Attack,Lin2022GCGRNN,Guo2024Simulation} so that our approach can be integrated with various applications of intelligent transportation systems.  
Lastly, the emergence of robot vehicles presents new opportunities for traffic control and optimization, especially under the context of mixed traffic where human-driven vehicles and robot vehicles coexist~\cite{Islam2024Heterogeneous,Wang2024Intersection,Wang2024Privacy,Poudel2024CARL,Poudel2024EnduRL,Villarreal2024Eco,Villarreal2023Pixel,Villarreal2023Chat}. 
We plan to explore how our technique can impact mixed traffic control and coordination in a meaningful way.   

\bibliographystyle{unsrt}
\bibliography{references}



\end{document}